\pdfoutput=1
\documentclass[sigconf,nonacm]{acmart}

\usepackage{multirow}
\usepackage{array}
\usepackage{tabularx}
\newcolumntype{Y}{>{\raggedright\arraybackslash}X}

\usepackage{amssymb}
\usepackage{pifont}
\usepackage{colortbl}
\usepackage{enumitem}
\usepackage{algorithm}
\usepackage[noend]{algpseudocode}
\definecolor{markyes}{HTML}{4F6B75}
\definecolor{markpart}{HTML}{A8812E}
\definecolor{markno}{HTML}{A0522D}
\definecolor{markna}{HTML}{8A8A8A}
\newcommand{\cmark}{\textcolor{markyes}{\ding{51}}}
\newcommand{\pmark}{\textcolor{markpart}{\ding{109}}}  %
\newcommand{\xmark}{\textcolor{markno}{\ding{55}}}
\newcommand{\nmark}{\textcolor{markna}{--}}
\newcommand{\cyes}[1]{\cmark~#1}
\newcommand{\cpart}[1]{\pmark~#1}
\newcommand{\cno}[1]{\xmark~#1}
\newcommand{\cna}[1]{\nmark~#1}

\begin{document}

\title{Testimony from Timelines: Evidence-Grounded Significance Testing of Temporal Clinical Hypotheses}

\author{Chengyang He}
\affiliation{%
  \institution{Stevens Institute of Technology}
  \city{Hoboken}
  \state{NJ}
  \country{USA}}
\email{che14@stevens.edu}

\author{Wenlong Zhang}
\affiliation{%
  \institution{Stevens Institute of Technology}
  \city{Hoboken}
  \state{NJ}
  \country{USA}}
\email{wzhang71@stevens.edu}

\author{Koduvayur Subbalakshmi}
\affiliation{%
  \institution{Stevens Institute of Technology}
  \city{Hoboken}
  \state{NJ}
  \country{USA}}
\email{ksubbala@stevens.edu}

\author{Ping Wang}
\authornote{Corresponding author.}
\affiliation{%
  \institution{Stevens Institute of Technology}
  \city{Hoboken}
  \state{NJ}
  \country{USA}}
\email{pwang44@stevens.edu}

\renewcommand{\shortauthors}{He et al.}

\begin{abstract}
Ask a language model whether sales tend to dip after a product recall, or
whether one side effect tends to follow another after a vaccine, and it
answers in a fluent sentence that reads like a finding, with no test behind
it. Assertions about how events unfold across many cases fill the Web, and each
is a temporal hypothesis that only a significance test over the right
records can settle. The verifier must specify the null, select the test and
decide which records bear on the hypothesis, and any of these choices can go
wrong while the conclusion still comes out right. Yet benchmarks for
language-model data analysis grade only the conclusion, with the relevant
data handed over. We study the problem in health, where a wrong verdict
is costly, and present \textsc{Strata}, a benchmark
that grades every decision on the way to a verdict: 8{,}735 hypotheses over the
clinical timelines of vaccine adverse-event reports, with gold plans,
evidence sets, $p$-values and verdicts derived under a released codebook and
reproduced by a deterministic oracle. We also propose \textsc{Art}
(articulate, retrieve, test), a staged paradigm that exposes each decision
as an inspectable output. Extensive experiments with five
model configurations show that systems fail mostly at choosing which records count, a
verdict--evidence gap that accuracy alone conceals. \textsc{Art} closes most of
this gap and brings a 31B open model to within 0.01 of the frontier models under the same paradigm; the frontier models succeed once given the codebook.
\end{abstract}



\maketitle

\brokenpenalty=0

\newcommand{\pend}[1]{\textit{#1}}
\newcommand{\sys}{\textsc{Strata}}
\newcommand{\art}{\textsc{Art}}

\section{Introduction}

\begin{figure}[t]
\centering
\includegraphics[width=0.9\columnwidth]{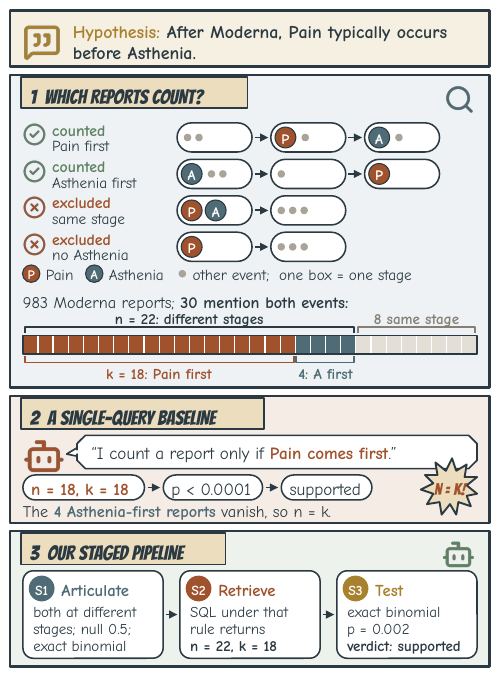}
\vspace{-10pt}
\caption{A worked example. (1) Four Moderna reports, one per category; the bar
sorts the 30 reports with both events into the numerator $N$ ($k = 18$), the
rest of the denominator $D$ ($n - k = 4$) and the 8 reports outside $D$, where
the two events share a stage. (2) A single-query baseline and (3) \art{}, with
the same open model.}
\label{fig:example}
\vspace{-10pt}
\end{figure}

\begin{figure*}[t]
\centering
\includegraphics[width=0.86\textwidth]{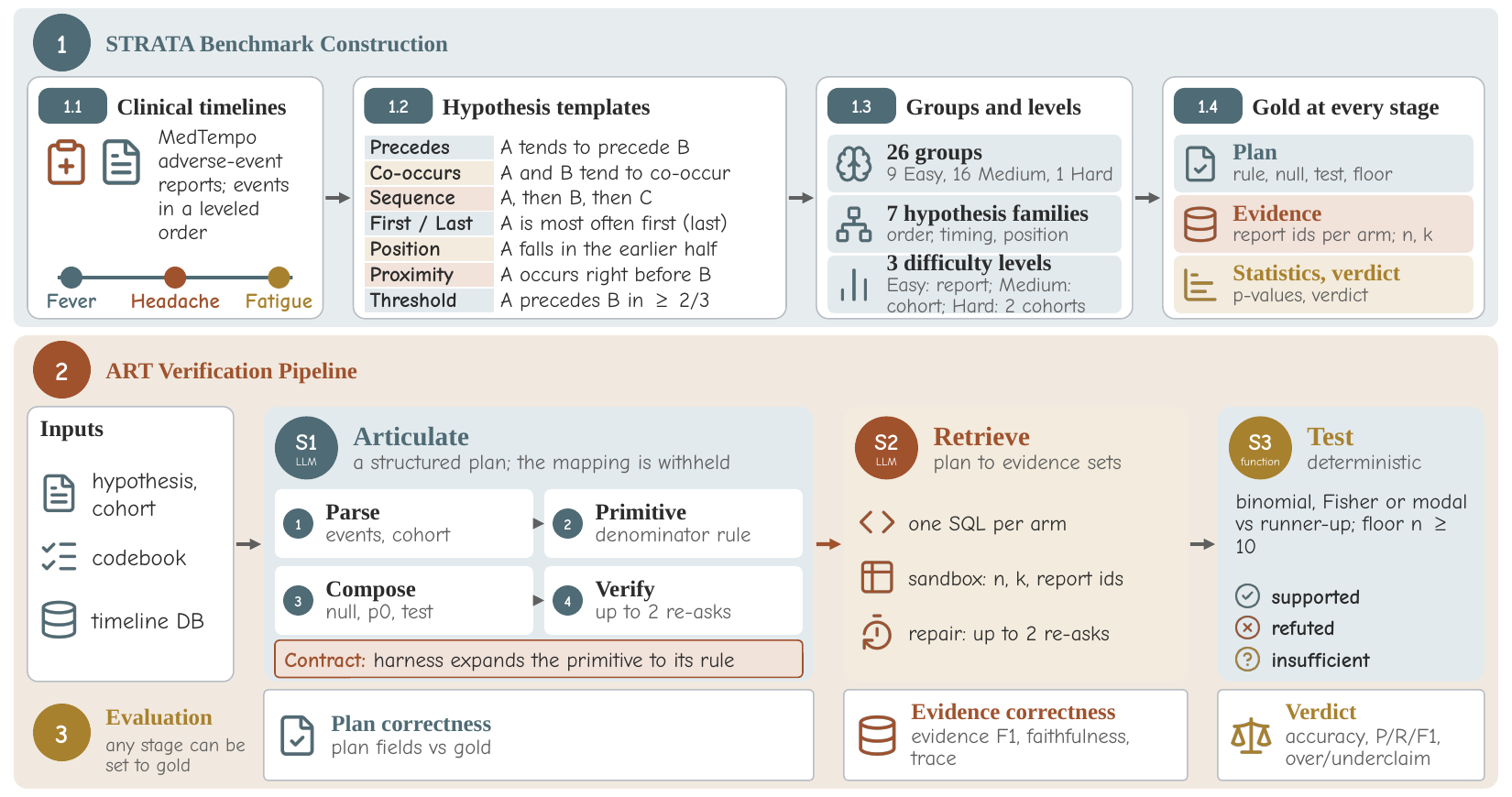}
\vspace{-12pt}
\caption{Overview of the \sys{} benchmark and the \art{} pipeline: (1)
construction, (2) verification, (3) the measure that scores each stage
against its gold.}
\label{fig:overview}
\vspace{-10pt}
\end{figure*}

\label{sec:intro}

Open a browser and the statements come at you. A post says that after a vaccine
one symptom tends to arrive before another. A summary says a treatment works
in most patients. An assistant answers a question over a public database in
a sentence that sounds settled. The Web has always held more such sentences
than anyone could check, and language models now write them, summarise them
and build the knowledge bases that answer them, in prose as confident as any
person's. Sorting the true ones from the false is the work of fact
verification, which checks each against a reference
\cite{allein2023implicit,barik2025chronofact,strong2025tsver}. But many of
the statements that matter are not facts waiting in a database. They are
hypotheses about a population, and in health the most consequential of them
are temporal: that one adverse event tends to follow another after a
vaccine, that a symptom comes first, that two events arrive together.
No single record settles a hypothesis like this, and the pattern a handful of
records seems to show may be chance. What settles it is a significance test, run over the right records
against the right null. Verifying such a hypothesis is \emph{temporal hypothesis verification}. Given a hypothesis about
the order or timing of events across a population of clinical timelines, a
system states the null, chooses the significance test, decides which
timelines can witness it, and returns \emph{supported},
\emph{refuted} or \emph{insufficient}.

A growing class of language-model systems promises exactly this
\cite{miao2026fisher,gim2026latch,majumder2025discoverybench}. Yet the
benchmarks that judge them \cite{miao2026fisher,zhu2024statqa,zhu2026statabench}
score the final answer with the relevant data handed over. The decisions
that make an answer statistical are never checked, and each of them can be
wrong while the verdict is right. Take the hypothesis \emph{``after Moderna, Pain
typically occurs before Asthenia''} (Figure~\ref{fig:example}). The null is
a 50/50 split of the reports in which the two events are ordered. The test
is a one-sided exact binomial at $\alpha = 0.05$ with a floor on the count. A
report where the two events share a stage witnesses neither order, so it
does not count. Each decision is a stipulation about what the sentence means
for this corpus, and the verdict does not show when one is wrong. In Figure~\ref{fig:example}, a single-query baseline counts only the reports
that support the hypothesis and still returns the right verdict: a
\emph{verdict--evidence gap}. A verdict that is right for the wrong reason
misleads where the tolerance for error is lowest, and existing benchmarks
hold no gold for the decisions behind it.

\sys{} (STatistical Reasoning Assessment for Temporal Assertions) is, to our
knowledge, the first benchmark for this task with gold for every decision:
the plan (null, test and conditioning rule), the evidence sets, the test
statistics and the verdict. It holds 8{,}735 hypotheses in 26 template
groups and four registers, generated over the MedTempo clinical timelines
of vaccine adverse-event reports \cite{craft2026}, with a held-out test set
of 1{,}500. Every instance carries its gold verdict (\emph{supported},
\emph{refuted} or a sub-typed \emph{insufficient}) and the gold of every
stage that produced it. All of it is computed mechanically from the
timelines and reproduced by a deterministic oracle. The stipulations the
labels depend on ship as a \emph{codebook}, written as text that any system
can be given. The timelines are the reference, as a knowledge base is for
fact verification, and the codebook prescribes the analysis rule.

We also introduce \art{} (articulate, retrieve, test), a staged
verification paradigm for the task. Its first stage articulates the hypothesis as
a plan, the second retrieves the evidence sets the plan names, and the third
runs the prescribed test deterministically. Each stage has an output
schema, and the two model stages have a bounded verify-and-repair loop. Any stage can be filled with
gold, and the prompts can state or withhold the codebook. The plan can reach
retrieval as prose or as a \emph{contract}, in which the harness expands
each primitive the plan names into its exact rule. These switches separate
two things that end-to-end evaluation conflates: the \emph{information} a
system is told and the \emph{architecture} it is given. Our contributions
are:
\begin{itemize}[leftmargin=*]
\item \textbf{A benchmark with gold at every stage of temporal hypothesis
verification}: \sys{}, whose cohort hypotheses about the order, adjacency, position and
co-occurrence of clinical events are each a significance decision, with the
null, test and population following from the temporal shape of each
hypothesis. Its 26
groups span seven families and three difficulty levels. Every instance ships its plan,
evidence sets, $p$-values and a three-decision verdict whose
\emph{insufficient} class is sub-typed.
\item \textbf{A generation process that makes the codebook part of the
task}: hypotheses are enumerated from the timelines by templates, and gold is
computed by the prescribed test for each hypothesis shape and validated by
permutation screens, invariants and an oracle. The rules that turn a
temporal sentence into a testable population ship as text.
\item \textbf{A staged verification paradigm}, \art{}, that makes every
decision an explicit, checkable output. It articulates a hypothesis as a plan
naming its conditioning rule, null and test, retrieves the conditioned
evidence sets under a contract resolved by the harness, and decides with the
prescribed test, run deterministically.
\end{itemize}

\section{Related Work}
\label{sec:related}

\textit{\textbf{Language models and statistical analysis}.} Work on
language models and statistical hypotheses ranges from choosing a
test to running a whole analysis. StatQA \cite{zhu2024statqa} asks which test applies,
StatFormBench \cite{wang2026statformbench} asks for the formulation of a
test family, and StatABench \cite{zhu2026statabench} scores the final
answer of an analysis run with fixed tool functions. P-Bench and Fisher-R1
\cite{miao2026fisher} are closest to our setting. An agent receives a
hypothesis with its null stated, a CSV of the relevant data and a glossary,
writes R to compute a $p$-value, and is scored on the decision and the
$p$-value. Agentic systems go further and plan the analysis themselves.
LATCH \cite{gim2026latch} runs LLM-chosen analyses through deterministic R
pipelines on EHR data with the method given. VERITAS \cite{veritas2026} is a
multi-agent co-scientist with a deterministic evidence classifier for
image-derived cohorts. DiscoveryBench and BLADE
\cite{majumder2025discoverybench,gu2024blade} evaluate open-ended discovery
agents against rubrics. Across this line of work, the data are either handed
over as the relevant subset or left entirely to the agent, the hypotheses
concern variables in a table, and only the final answer or a rubric is
scored. None asks the system to decide which records a statistic is computed
over, and none releases gold for that decision or for the null and the test
around it.

\begin{table}[t]
\centering
\footnotesize
\setlength{\tabcolsep}{3.6pt}
\begin{tabular}{@{}lcccccc@{}}
\toprule
Work & Null & \shortstack{Test\\choice} & \shortstack{Test\\run} & \shortstack{Temporal\\hypotheses} & Clinical & \shortstack{Stage\\gold} \\
\midrule
StatQA \cite{zhu2024statqa} & \pmark & \cmark & \xmark & \xmark & \xmark & \xmark \\
StatFormBench \cite{wang2026statformbench} & \pmark & \pmark & \xmark & \xmark & \xmark & \pmark \\
StatABench \cite{zhu2026statabench} & \pmark & \cmark & \pmark & \xmark & \xmark & \xmark \\
P-Bench \cite{miao2026fisher} & \xmark & \pmark & \pmark & \pmark & \pmark & \pmark \\
DiscoveryBench \cite{majumder2025discoverybench} & \pmark & \pmark & \pmark & \xmark & \xmark & \pmark \\
EHRSQL \cite{lee2022ehrsql} & \xmark & \xmark & \xmark & \xmark & \cmark & \pmark \\
EHRAgent \cite{shi2024ehragent} & \xmark & \xmark & \xmark & \xmark & \cmark & \nmark \\
LATCH \cite{gim2026latch} & \xmark & \xmark & \cmark & \xmark & \cmark & \nmark \\
VERITAS \cite{veritas2026} & \pmark & \pmark & \pmark & \xmark & \cmark & \nmark \\
\midrule
\textbf{\sys{} + \art{}} & \cmark & \cmark & \cmark & \cmark & \cmark & \cmark \\
\bottomrule
\end{tabular}
\caption{Comparison of representative related work on what each work asks and scores (full version:
Appendix~\ref{app:positioning}). \emph{Test choice}: chosen and scored;
\emph{Test run}: the verdict follows from the test result; \emph{Stage gold}: gold
beyond the answer. \cmark{} yes, \pmark{} partly, \xmark{} no, \nmark{} n/a.}
\label{tab:related-main}
\end{table}

\textit{\textbf{Temporal reasoning and clinical records}.} Temporal
reasoning with language models has mostly been evaluated
on single texts. Benchmarks ask about the order, duration or frequency of
events in a passage or a synthetic event graph
\cite{wang2024tram,fatemi2025testoftime}, or in real-world text at several
levels \cite{wei2025time,luo2025etrqa}. Temporal fact verification judges
one statement against its retrieved evidence, grounding its time expressions
\cite{allein2023implicit,barik2024time}, ordering its events on a timeline
\cite{barik2025chronofact} or reading a time series \cite{strong2025tsver}.
In the clinical domain, temporal
relation extraction labels the order of events within one narrative
\cite{sun2013i2b2,bethard2016clinicaltempeval}, and TIMER \cite{timer2025}
and CRAFT \cite{craft2026} extract and refine clinical timelines with
language models. Question answering over clinical databases likewise asks
for one answer per question: EHRSQL \cite{lee2022ehrsql} adds an abstention
option, EHRAgent \cite{shi2024ehragent} a code-writing agent with
error-driven retries, and open text-to-SQL models such as OmniSQL
\cite{li2025omnisql} are built for the same setting. In all of these, the
answer is a fact about one text, one series, one timeline or one query. A
temporal hypothesis is an assertion about a population of timelines, and
verifying it adds the decisions this paper scores: which timelines can
witness the order at all, what chance looks like when events crowd the
early stages, and which test decides. \sys{} brings the two lines together: it poses temporal hypotheses as
significance tests over the MedTempo timelines released with CRAFT, and
EHRSQL, EHRAgent and DiscoveryBench are adapted to it as baselines.
Table~\ref{tab:related-main} compares the closest work, and
Appendix~\ref{app:positioning} gives the full comparison.

\section{The \sys{} Benchmark}
\label{sec:dataset}

\subsection{Task formulation}
\label{sec:task}

The input is a hypothesis $h$ and the timelines of a report set
$\mathcal{R}$. Each report's timeline is a \emph{leveled order}: a chain of
\emph{stages}, each holding the MedDRA-coded events \cite{brown1999meddra} reported at that
point; a stage is a unit of order in the annotation, not a calendar day.
A term's first mention fixes its stage $s_r(e)$ in report $r$, and events at
one stage \emph{co-exist}, with no order between them. The reports of one
vaccine form a \emph{cohort} $\mathcal{C} \subseteq \mathcal{R}$. The
hypothesis is stated in natural language in one of four registers and has
one of three difficulty levels: an \emph{Easy} hypothesis concerns one named
report and is read off its timeline with no test, a \emph{Medium} one is
tested within one cohort, and a \emph{Hard} one compares two cohorts. The system sees each report's
vaccine and stages, never its text. A supported hypothesis describes a regularity in what was
reported to VAERS, not a causal effect.

A cohort hypothesis asserts a rate over the reports that can witness it.
Its \emph{denominator} $D \subseteq \mathcal{C}$ is that set, and its \emph{numerator} $N \subseteq D$ is the
subset that supports it. For ``$A$ precedes $B$'',
$D = \{r \in \mathcal{C} : A, B \in r,\ s_r(A) \neq s_r(B)\}$ and
$N = \{r \in D : s_r(A) < s_r(B)\}$; for ``$A$ and $B$ co-occur'',
$D = \{r \in \mathcal{C} : A, B \in r\}$ and $N = \{r \in D : s_r(A) = s_r(B)\}$,
the same-stage reports. The pair $(D, N)$ forms an \emph{arm}
with counts $n = |D|$ and $k = |N|$. Most hypotheses have one arm and
compare $k/n$ with a \emph{null} $p_0$, the rate expected by chance
($p_0 = 0.5$ for an order hypothesis). A two-arm hypothesis tests whether the
rates of its arms are equal; the arms are two event pairs in one cohort,
the reports of one cohort with and without an event $A$, compared on their rate of $B$ (both Medium), or one pair in two cohorts (Hard).

Given $h$ and the timelines, a system returns three outputs. The
\emph{plan} $\pi$ names the quantity compared, the null, the tail, the test
family, the number of arms $K \in \{1, 2\}$ and the \emph{denominator
primitive}, the rule that maps the events and cohort of the hypothesis to
$D$. The
\emph{evidence} $E = \{(D_j, N_j)\}_{j=1}^{K}$ lists the reports in each
arm; for \emph{first/last} (argmax) hypotheses each report in $D_j$ also
carries the candidate it witnesses, so $E$ partitions $D_j$ by candidate and
$N_j$ is the named candidate's cell. The \emph{verdict} $y$ follows from the test's one-sided $p$-values,
$p^{+}$ in the asserted direction and $p^{-}$ against it. If some arm has
fewer than $n_{\min} = 10$ reports (the \emph{floor}), $y$ is \emph{insufficient}:
\emph{not-locatable} when some arm is empty and \emph{sparse} otherwise.
Otherwise $y$ is \emph{supported} if $p^{+} \le \alpha$, \emph{refuted} if
$p^{-} \le \alpha$ ($\alpha = 0.05$), and \emph{insufficient}
(\emph{inconclusive}) if neither; a supported or refuted verdict is
\emph{decisive}. Since $\alpha$ applies to each direction, a decisive
verdict has probability at most 0.10 under the null, and 0.045 on
average on the test set.

Each output is scored against its own gold, $\pi^{*}$, $E^{*}$ and $y^{*}$;
the gold verdict also carries its sub-type and both $p$-values. At several points the reading of a hypothesis depends on a stipulation about the corpus, not on its wording: whether a report with both events at one stage counts for
``precedes'', or whether a tie for ``first'' leaves the denominator. The
benchmark ships these stipulations as the codebook $\mathcal{B}$
(Section~\ref{sec:frames}) and withholds only the \emph{mapping}
$h \mapsto \pi$ from wording to rule and test; composing that mapping is what a system is scored on.

\subsection{Benchmark construction}
\label{sec:construction}
\label{sec:corpus}

\textit{\textbf{Source timelines}.} We use the clinical timelines of
MedTempo~\cite{craft2026}: 3{,}166 VAERS
adverse-event reports \cite{shimabukuro2015vaers} for three COVID-19 vaccines (1{,}164 Janssen, 1{,}019
Pfizer, 983 Moderna), with 1{,}660 event terms and on average 7.2 events over
3.4 stages per report.

\begin{table}[t]
\centering
\small
\setlength{\tabcolsep}{4pt}
\begin{tabular}{@{}lr@{}}
\toprule
Test set (Medium / Hard) & 1{,}500 (1{,}418 / 82) \\
Distinct facts & 861 \\
Single-report case study (Easy) & 180 \\
Development set (4--10 hypotheses per group) & 184 \\
Groups / families / difficulty levels & 26 / 7 / 3 \\
Registers per hypothesis & 4 \\
Pool: hypotheses (texts) & 8{,}735 (34{,}940) \\
Easy / Medium / Hard & 3{,}600 / 5{,}048 / 87 \\
Supported / Refuted / Insufficient & 3{,}459 / 3{,}455 / 1{,}821 \\
Inconclusive / Sparse / Not-locatable & 712 / 495 / 614 \\
Denominator size $n$, median (IQR) & 18 (12--34) \\
\bottomrule
\end{tabular}
\caption{Basic data statistics. Per-group sizes are in the hypothesis catalogue
(Appendix~\ref{app:catalogue}); Easy hypotheses have no insufficient class.}
\label{tab:stats}
\vspace{-20pt}
\end{table}

\label{sec:space}\label{sec:realization}
\textit{\textbf{Hypotheses}.} The hypotheses come from 26 template groups in seven hypothesis families:
\emph{precedes}, \emph{co-occurs}, \emph{sequence} of three events,
\emph{first/last} (argmax) among candidates, and, to diversify the null,
\emph{position} (an event in the earlier half of a timeline),
\emph{proximity} (two events at consecutive stages) and \emph{threshold} (a
rate of at least a stated bar). The groups yield a pool of 8{,}735
hypotheses: 3{,}600 Easy (9 groups), 5{,}048 Medium (16) and 87 Hard (1).
Hard hypotheses are rare because few event pairs are
testable in two cohorts at once, and sequence appears at the Easy level
only.
Each group is rendered in four registers (\textsc{Original},
\textsc{Research}, \textsc{Media}, \textsc{Clinical}) by mechanical
slot-filling (Table~\ref{tab:catalogue}). Each hypothesis is built on a \emph{core}:
its template group, events and cohort, hence its denominator primitive,
tested quantity, null and arm structure, which fix the statistic whatever
the wording. A core is \emph{decisive} when its test is significant
in one direction. Every decisive core is posed both ways, as a
\emph{supported} hypothesis and its \emph{mirror} in the opposite direction, so a
system that supports both directions of one fact is caught. Each cohort
group ships near-equal thirds of the three verdicts (C-V3 and Y-V3 ship more
\emph{insufficient}, Table~\ref{tab:catalogue}), so its wording cannot
predict the label. \emph{Not-locatable} hypotheses use real cohort vocabulary whose
joint count is zero, so only querying the timelines reveals the empty
denominator.

\label{sec:gold}
\textit{\textbf{Gold labels}.} Labels are computed mechanically from the timelines. The null follows
from the shape of the hypothesis: 0.5 for order hypotheses, the named cell against its
strongest competitor for argmax hypotheses, equality of two rates for
two-arm hypotheses, a bar stated in the wording for threshold hypotheses, and a rate
derived from the data for position and proximity hypotheses. Order and bar
hypotheses use a one-sided exact binomial, and data-derived nulls a binomial at
the mean per-report chance, a conservative bound on the exact
Poisson-binomial test. Argmax hypotheses use an exact binomial on that comparison: an intersection--union test in the asserted direction
and, in the opposite direction, a Bonferroni-adjusted union over the
competitors (Appendix~\ref{app:testdetails}); two-arm
hypotheses use Fisher's exact test
\cite{fisher1935design}, extended to reports that sit in both arms. Every test has size at most $\alpha$ per direction. Which reports count is the substance of
the task: \emph{precedes} counts only reports with both events at distinct
stages, \emph{co-occurs} admits every report with both and counts the same-stage ones (one co-occurs group splits the cohort on $A$ and counts $B$ anywhere), and \emph{first/last} drops
reports in which candidates tie. The only sample-size gate is the floor of $n_{\min} = 10$ reports per
arm; it defines the \emph{not-locatable} and \emph{sparse} sub-types, while \emph{inconclusive} comes from the test. Each cohort hypothesis releases both one-sided $p$-values, the Wilson
interval \cite{wilson1927} and Cohen's effect size \cite{cohen1988}, oriented to
the hypothesis as asserted.

\label{sec:screens}
\textit{\textbf{Validation}.} Before shipping, every group cleared a permutation null matched to its
hypothesis shape \cite{good2005permutation}, under which decisive labels fall
to about $\alpha$ of the cores (Appendix~\ref{app:harness}); six families
of invariants hold over all
51{,}855 instances enumerated before the per-group cap, and a deterministic oracle that parses each
template back into the reference SQL reproduces every gold label.

\label{sec:release}\label{sec:stats}
\textit{\textbf{Release}.} The benchmark ships (Table~\ref{tab:stats}) a 1{,}500-hypothesis
test set, drawn
as 61--96 hypotheses per cohort group, balanced across the verdicts (484 / 483
/ 533) and fact-disjoint from a development set that draws 4--10 hypotheses
from every group; a 180-hypothesis
single-report case study; and the full pool of 8{,}735 hypotheses with its
generator. Hypotheses that pose the same statistic over the same events and cohort,
such as a hypothesis and its mirror or one hypothesis in two answer formats,
share a \texttt{fact\_id} (a bar hypothesis and an order hypothesis on the same counts do not, since their nulls differ); the 1{,}500 test hypotheses hold 861 facts, the
unit of the split and of every interval we report.
Report identifiers are pseudonymous, and the full release, with the code,
follows acceptance.

\section{The Proposed Method}
\label{sec:framework}
\label{sec:method}
\label{sec:contracts}
\label{sec:verify}

\begin{table*}[t]
\centering
\footnotesize
\setlength{\tabcolsep}{2.2pt}
\renewcommand{\arraystretch}{1.0}
\begin{tabular}{@{}lrrrr@{\hspace{6pt}}rrrr@{\hspace{6pt}}rrrr@{\hspace{6pt}}rrrr@{\hspace{6pt}}rrrr@{}}
\toprule
Methods & \multicolumn{4}{c}{\textbf{Qwen2.5-32B pair}} & \multicolumn{4}{c}{\textbf{qwen3.8-27B}} & \multicolumn{4}{c}{\textbf{gemma4-31B}} & \multicolumn{4}{c}{\textbf{Opus 5.5}} & \multicolumn{4}{c}{\textbf{GPT-5.6}} \\
\cmidrule(lr){2-5}\cmidrule(lr){6-9}\cmidrule(lr){10-13}\cmidrule(lr){14-17}\cmidrule(lr){18-21}
 & Acc & P & R & F1 & Acc & P & R & F1 & Acc & P & R & F1 & Acc & P & R & F1 & Acc & P & R & F1 \\
\midrule
EHRSQL & .528 & .552 & .522 & .516 & .593 & .611 & .588 & .585 & .634 & .647 & .630 & .627 & .867 & .877 & .866 & .868 & .872 & .872 & .871 & .871 \\
EHRSQL-C & .577 & .606 & .571 & .571 & .744 & .763 & .740 & .743 & \underline{.857} & \underline{.887} & \underline{.854} & \underline{.860} & \textbf{1.000} & \textbf{1.000} & \textbf{1.000} & \textbf{1.000} & \textbf{.984} & \textbf{.984} & \textbf{.983} & \textbf{.983} \\
EHRAgent & \underline{.596} & \underline{.689} & \underline{.602} & \underline{.592} & \underline{.838} & \underline{.858} & \underline{.836} & \underline{.842} & .733 & .738 & .732 & .734 & \multicolumn{4}{c}{---} & .796 & .809 & .795 & .798 \\
DiscoveryBench & \multicolumn{4}{c}{n/a} & .650 & .823 & .647 & .716 & .674 & .763 & .671 & .706 & \multicolumn{4}{c}{---} & \multicolumn{4}{c}{---} \\
\midrule

\textbf{\art{} (Ours)} & \textbf{.806} & \textbf{.878} & \textbf{.801} & \textbf{.822} & \textbf{.869} & \textbf{.893} & \textbf{.867} & \textbf{.878} & \textbf{.981} & \textbf{.983} & \textbf{.981} & \textbf{.982} & \underline{.991} & \underline{.991} & \underline{.990} & \underline{.991} & \underline{.981} & \underline{.982} & \underline{.981} & \underline{.981} \\
\quad \art{} {\scriptsize w/o} verify-and-repair & .757 & .847 & .751 & .771 & .727 & .804 & .720 & .734 & .837 & .879 & .833 & .842 & .999 & .999 & .999 & .999 & .999 & .999 & .999 & .999 \\
\quad \art{} {\scriptsize w/o} harness test (model verdict) & .517 & .646 & .506 & .484 & .577 & .615 & .568 & .509 & .779 & .812 & .774 & .777 & \multicolumn{4}{c}{---} & \multicolumn{4}{c}{---} \\
\quad \art{} {\scriptsize w/o} contract (plan as prose) & .663 & .703 & .657 & .668 & .823 & .857 & .819 & .831 & .947 & .949 & .946 & .947 & \multicolumn{4}{c}{---} & \multicolumn{4}{c}{---} \\
\bottomrule
\end{tabular}
\caption{Main results on the test set: verdict accuracy (Acc) and macro P, R
and F1 over the three verdicts; leading zeros dropped. Rows below \art{}
remove one of its components. Under the Qwen2.5 pair, the single-query
baselines and the agents run Qwen2.5-Coder alone. 95\% intervals on Acc are
at most $\pm 0.03$. Bold/underline: best/second best among the adapted
baselines and \art{}. Dash: run not made, to bound API cost; n/a: the agent
does not run with this model.}
\label{tab:main}
\end{table*}

\begin{figure*}[t]
\centering
\includegraphics[width=\textwidth]{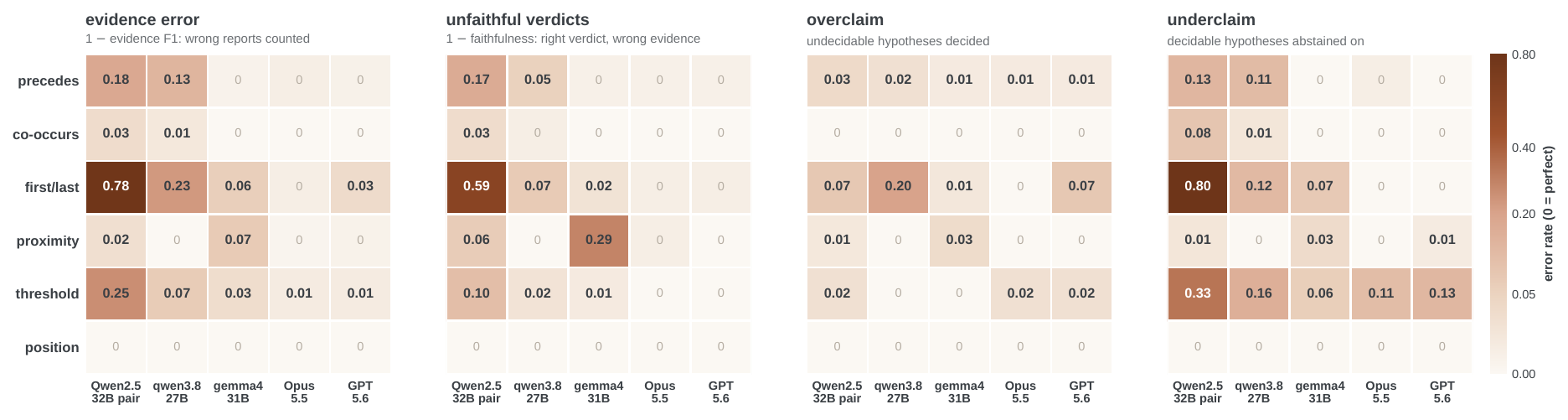}
\vspace{-15pt}
\caption{Evidence errors of \art{} on the test set, by hypothesis family and
model (Section~\ref{sec:metrics}); 0 is perfect in every panel. The colour
scale is square-root.}
\label{fig:families}
\vspace{-10pt}
\end{figure*}

\begin{algorithm}[t]
\caption{\art{} on one cohort hypothesis. S1 and S2 call the model, each with up to $T$ re-asks ($T = 2$ in every reported \art{} row); the checks and S3 are deterministic; $\rho$ is the resolved contract, $V$ and $W$ what the checks flag.}
\label{alg:art}
\small
\begin{algorithmic}[1]
\Require hypothesis $h$, timelines $\mathcal{R}$, codebook $\mathcal{B}$, re-ask cap $T$
\Ensure plan $\pi$, evidence $E$, verdict $y$
\State $\pi \gets \textsc{Articulate}(h, \mathcal{B})$ \Comment{S1}
\For{$t \gets 1$ \textbf{to} $T$} \Comment{plan repair}
  \State $V \gets \textsc{PlanChecks}(\pi, h, \mathcal{B})$
  \If{$V = \emptyset$} \textbf{break}
  \EndIf
  \State $\pi \gets \textsc{Articulate}(h, \mathcal{B}, V)$
\EndFor
\If{$\pi$ names no known hypothesis type} \Return $(\pi, \emptyset, \textit{insufficient})$
\EndIf
\State $\rho \gets \textsc{Contract}(\pi)$ \Comment{rule of $\pi$'s primitive}
\State $E \gets \textsc{Execute}(\textsc{Retrieve}(\pi, \rho), \mathcal{R})$ \Comment{S2}
\For{$t \gets 1$ \textbf{to} $T$} \Comment{retrieval repair}
  \State $W \gets \textsc{RowChecks}(E, \pi)$
  \If{$W = \emptyset$} \textbf{break}
  \EndIf
  \State $E \gets \textsc{Execute}(\textsc{Retrieve}(\pi, \rho, W), \mathcal{R})$
\EndFor
\State $(p^{+}, p^{-}) \gets \textsc{Test}(\pi, E)$ \Comment{S3}
\State $y \gets \textsc{Verdict}(p^{+}, p^{-}, \min_j |D_j|)$ \Comment{Section~\ref{sec:task}}
\State \Return $(\pi, E, y)$
\end{algorithmic}
\end{algorithm}

\art{} produces the scored outputs of Section~\ref{sec:task} in three
stages (Figure~\ref{fig:overview}, Algorithm~\ref{alg:art}): it
\emph{articulates} the hypothesis as a plan naming its null, test and
conditioning rule, \emph{retrieves} the evidence sets the plan conditions on,
and \emph{tests} them with the prescribed test, run deterministically, so
the verdict is a statistical decision rather than a model judgment. Each
stage has an output schema and can be scored against its gold or filled
with gold. The two model stages have a bounded verify-and-repair loop that
reads only the contract, the codebook and the data, never the gold; unless
stated otherwise, \art{} includes these loops.

\subsection{S1: Articulate the plan}
\label{sec:frames}

From the hypothesis text, its difficulty level and its cohort, the model composes a plan:
\texttt{events}, \texttt{quantity}, \texttt{denominator} (a named primitive
such as \emph{ordered\_pair} or \emph{all\_present\_unique\_first}), \texttt{arms}, \texttt{null\_kind} with $p_0$
or a reference population, \texttt{tail}, \texttt{test\_family} with its
rule, and the floor. The prompt states the codebook, organised by situation, with a computed dataset card, the null sources and the test definitions
(Appendix~\ref{app:harness}). It withholds the mapping from wording to hypothesis type, so the model composes
the plan instead of looking it up; the prompt is zero-shot, since a worked
plan would be a row of that mapping, and the harness derives the hypothesis type
from the plan by reverse lookup over its fields. A
plan that names no known hypothesis type is scored as
insufficient (Appendix~\ref{app:harness}).
The plan is then checked for consistency with the codebook; on a violation the model is re-asked with the violations named, up to two re-asks.

\begin{figure*}[t]
\centering
\includegraphics[width=\textwidth]{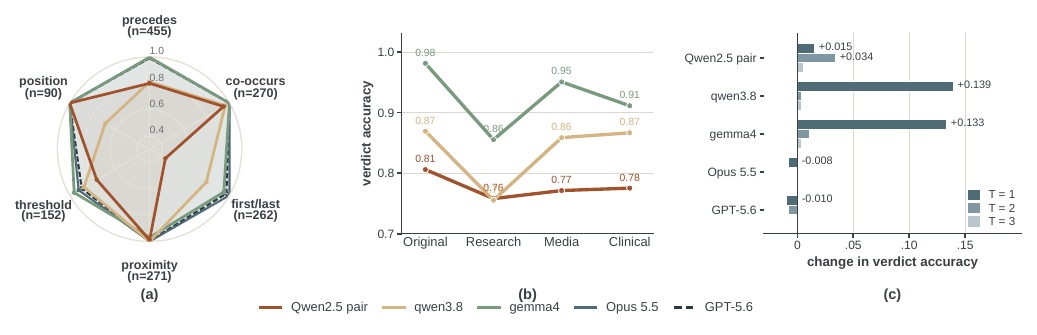}
\vspace{-25pt}
\caption{\art{} on the test set: (a) verdict accuracy by hypothesis family; (b)
accuracy by paraphrase register, open models; (c) change in verdict
accuracy when the cap on re-asks per stage rises from $T-1$ to $T$ (\art{} as reported uses $T = 2$).}
\label{fig:panels}
\end{figure*}

\subsection{S2: Retrieve the evidence sets}

For each arm the model writes one SQL statement over the timeline database
(two tables, three views) that returns \emph{one row per report} with three Boolean flags (in the denominator, in the numerator, excluded) and an optional cell label for argmax hypotheses. The plan reaches this stage as a \emph{contract}.
The harness resolves the plan's hypothesis type and writes that type's
conditioning rule and arm plan into the prompt, so the model receives the
rule for the one primitive the plan names. Without the contract, S2 receives the plan's JSON with the full codebook
as \emph{prose} (Section~\ref{sec:modular}). The harness executes the
statement in a read-only sandbox and aggregates the rows into $(n, k)$ per arm
with the identifier sets, so the evidence is auditable against the gold
sets. A rejected statement or rows that violate the plan's primitive are
violations; an arm whose denominator equals its numerator, or an empty arm beside a full one, triggers a heuristic recheck, since both can be correct. Either triggers a re-ask, up to two re-asks, and the last answer replaces the earlier ones, so a correct statement that trips a recheck is kept only if the model
returns it again. The contract and the checks are written
by hand for this corpus and defined against the plan
(Appendix~\ref{app:harness}).

\subsection{S3: Test the hypothesis}

A deterministic function runs the plan's test on the counts
(Table~\ref{tab:catalogue}): the binomial against the plan's $p_0$ (for
a data-derived null, $p_0$ is the mean chance over the retrieved
denominator), Fisher's test for two arms (with shared reports exchanged,
Appendix~\ref{app:testdetails}), and, for argmax hypotheses, the named cell against its strongest competitor (Section~\ref{sec:gold}); the codebook also admits the asymptotic two-proportion test for large
cells. S3 applies the floor and returns the verdict with both tails.
Nothing is inferred at this stage, so it needs no verification.

\section{Experiments}
\label{sec:experiments}

\begin{table}[t]
\centering
\footnotesize
\setlength{\tabcolsep}{3.2pt}
\begin{tabular}{@{}llrrrrr@{}}
\toprule
 & & Qwen2.5 & qwen3.8 & gemma4 & Opus & GPT \\
Stage & Methods & 32B pair & 27B & 31B & 5.5 & 5.6 \\
\midrule
\multirow{3}{*}{S1} & \art{}: Articulate (Ours) & \textbf{.952} & \textbf{.915} & \textbf{.992} & .991 & .991 \\
 & Plain prompt & .871 & .856 & .919 & --- & --- \\
 & Plain prompt + CoT & .923 & .859 & .909 & --- & --- \\
\midrule
\multirow{2}{*}{S2} & \art{}: Retrieve (Ours) & \textbf{.833} & .943 & .989 & 1.000 & .991 \\
 & OmniSQL-32B$^\ddagger$ & .740 & n/a & n/a & n/a & n/a \\
\midrule
\multirow{2}{*}{S3} & \art{}: Test (Ours) & \textbf{1.000} & \textbf{1.000} & \textbf{1.000} & \textbf{1.000} & \textbf{1.000} \\
 & StatQA & .620 & .628 & .787 & .980 & .981 \\
\bottomrule
\end{tabular}
\caption{Stage baselines on the test set: verdict accuracy with one stage
filled by \art{}'s component or a baseline, the others gold or deterministic.
CoT: zero-shot chain-of-thought. $^\ddagger$Fine-tuned from Qwen2.5-Coder-32B.
Bold: best per column within a stage where a baseline exists. Dash: run not
made.}
\label{tab:stage}
\vspace{-25pt}
\end{table}

\subsection{Research questions}
\label{sec:rqs}

The experiments ask whether systems make the decisions of
Section~\ref{sec:task} correctly and whether their verdicts show it. We
first ask how well current systems verify these temporal hypotheses end to
end, and how the codebook, the model and the staged design each change
accuracy (\textbf{RQ1}). We then ask where \art{}'s gain comes from and
where its accuracy still varies: by stage, hypothesis family, paraphrase
register and difficulty level (\textbf{RQ2}). Last, we ask whether a correct
verdict rests on correct evidence, and whether systems decide exactly the
hypotheses the data can decide (\textbf{RQ3}).
\vspace{-2.5mm}

\subsection{Baselines}
\label{sec:baselines}

Three published systems are adapted to the benchmark, all over the same
database. \emph{EHRSQL} \cite{lee2022ehrsql} follows the single-query
text-to-SQL protocol: one prompt per hypothesis asks for one query over our schema
that marks each report as in the denominator, the numerator or neither, and
the harness runs the query and the test. \emph{EHRSQL-C} adds the
codebook to the prompt, so it differs from \art{} only in having no plan, no
contract and no repair. \emph{EHRAgent} \cite{shi2024ehragent} runs as
released, writing and executing its own code with up to 12 error-driven retries and four demonstrations, and runs its own test. It receives the
schema, the test functions and the floor but not the codebook, since its
knowledge stage writes the plan itself. \emph{DiscoveryBench}'s reference
agent \cite{majumder2025discoverybench} runs on the same tables
exported as CSV, with the codebook as its domain knowledge, an instruction to
end on a verdict line and up to 25 steps. The agents measure autonomous analysis; EHRSQL-C is the matched comparison.

\emph{Stage baselines} replace one stage of \art{} with an established
method and leave the other stages gold or deterministic, so each row
measures the loss from that stage alone (Table~\ref{tab:stage}): for
Articulate, a plain prompt with the dataset description, codebook and plan
schema but without \art{}'s explanations, field rules or verify-and-repair loop,
with and without zero-shot chain-of-thought \cite{kojima2022zeroshot}; for
Retrieve, OmniSQL \cite{li2025omnisql}, fine-tuned from the base of the
Qwen2.5 pair's SQL model, writing the SQL from the gold plan; for Test, the
StatQA protocol \cite{zhu2024statqa}, which gives the model the gold counts
and test definitions and asks for the test and the verdict. The matching
\art{} rows fill the same stage with \art{}'s own component.

\subsection{Evaluation metrics}
\label{sec:metrics}

The verdict is scored as a three-class prediction, by accuracy and by
precision, recall and F1 macro-averaged over \emph{supported},
\emph{refuted} and \emph{insufficient}. Five evidence measures use the stage gold to score the steps behind a
verdict.
\emph{evidence F1} is the F1 between the reports a system placed in an arm's
denominator and the gold set, averaged over arms. \emph{faithfulness} (of the denominator) is the share of correct verdicts whose denominator is exactly
the gold set in every arm. \emph{overclaim} is the share of gold-\emph{insufficient} hypotheses
given a decisive verdict, and \emph{underclaim} is the share of decisive
hypotheses answered \emph{insufficient}. \emph{trace} is the share of hypotheses on which the decisions behind the verdict are jointly right: both evidence
sets in every arm, the test family, the null kind and the verdict; the
null's value $p_0$, the tail and the floor are scored as plan fields and are not part of trace. The trace score extends the FEVER
score of fact verification, which credits a verdict only when its evidence
is also correct \cite{thorne2018fever}, to the whole chain of decisions.

\subsection{Experimental settings}
\label{sec:setup}

Unless stated otherwise, results are on the 1{,}500-hypothesis test set in
the \textsc{Original} register, run once after every prompt was frozen on the
development set; the single-report study (Table~\ref{tab:report}) and the
register reruns (Figure~\ref{fig:panels}b) are the exceptions. A review copy (development set with stage gold, timeline database,
outputs, scorer) is at
\url{https://anonymous.4open.science/r/STRATA-WWW2027/}. Five model configurations fill the model slots: the Qwen2.5 pair (Qwen2.5-32B-Instruct for planning, Qwen2.5-Coder-32B for SQL \cite{qwen2025qwen25}); two further open models, qwen3.8-27B \cite{qwen2026qwen38} and gemma4-31B \cite{gemma2026gemma4} (qwen3.8 and gemma4 below), each filling both stages, thinking off; and the frontier models Claude Opus~5.5
\cite{anthropic2026opus55} and GPT-5.6 \cite{openai2026gpt56}. Open models run 4-bit under Ollama on one RTX A5000; frontier models run
over their APIs (Appendix~\ref{app:harness}). The three open models run every stage baseline and ablation, and every system that runs with them. The frontier models run EHRSQL, EHRSQL-C, \art{}, EHRAgent (GPT-5.6), the stage
rows of \art{}, StatQA and the verify-and-repair ablation; the remaining
frontier cells and the register reruns, where the single query already
scores 0.98--1.00, are not run, to bound API cost. Verdict
accuracies carry 95\% bootstrap intervals
\cite{efron1993bootstrap} over facts, and paired comparisons use a
fact-level sign-flip test (Appendix~\ref{app:harness}). Cost and latency per hypothesis are in Appendix~\ref{app:cost}. All experiments, development
runs included, took about 355 GPU-hours and \$246 in API fees.

\section{Results}
\label{sec:results}

\begin{figure}[t]
\centering
\includegraphics[width=0.92\columnwidth]{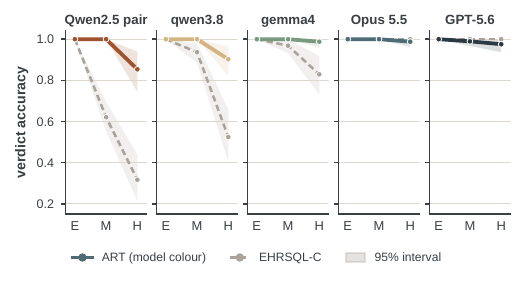}
\vspace{-10pt}
\caption{Verdict accuracy on one hypothesis shape, \emph{A precedes B}, by
difficulty: Easy, one report (P-R1, 20 hypotheses); Medium, one cohort (P-V1, 95);
Hard, two cohorts (P-M2, 82). One panel per model, \art{} against EHRSQL-C;
bands are 95\% bootstrap intervals over facts.}
\label{fig:tiers}
\vspace{-10pt}
\end{figure}

\begin{table}[t]
\centering
\footnotesize
\setlength{\tabcolsep}{4pt}
\renewcommand{\arraystretch}{0.95}
\begin{tabular}{@{}lrrrrr@{}}
\toprule
 & Qwen2.5 & qwen3.8 & gemma4 & Opus & GPT \\
Methods & 32B pair & 27B & 31B & 5.5 & 5.6 \\
\midrule
EHRSQL & .433 & .989 & .744 & --- & --- \\
EHRSQL-C & .750 & \textbf{1.000} & \textbf{1.000} & \textbf{1.000} & \textbf{1.000} \\
\textbf{\art{}} & \textbf{1.000} & \textbf{1.000} & \textbf{1.000} & .983 & \textbf{1.000} \\
\bottomrule
\end{tabular}
\caption{Verdict accuracy on the 180 single-report hypotheses (20 per
group). Bold: best per model; a dash marks a run not made.}
\label{tab:report}
\vspace{-25pt}
\end{table}

\subsection{RQ1: end-to-end verification}
\label{sec:main}

\art{} is the most accurate end-to-end system for every open model
(Table~\ref{tab:main}): it gains $+0.23$ over EHRSQL-C for the Qwen2.5
pair, $+0.13$ for qwen3.8 and $+0.12$ for gemma4, each $p < 0.001$ by the
paired sign-flip test over facts, and gemma4 under \art{} (0.981) sits
within 0.01 of \art{} with the frontier models, at no marginal API fee.

Accuracy depends first on whether the system is told the codebook: EHRSQL-C
gains over EHRSQL 0.05 for the Qwen2.5 pair, 0.15 for qwen3.8, 0.22 for
gemma4 and 0.11--0.13 for the frontier models, which then verify 0.98--1.00
of the hypotheses. The frontier models lack only the corpus
rules, which no wording states. The Qwen2.5 pair cannot apply them from
text alone; the staged design applies them for it (0.806 against 0.577).

At the frontier the single query already solves the verdict ($-0.009$,
$p < 0.001$, and $-0.003$, $p = 0.70$, for \art{} against EHRSQL-C), and
\art{} adds an auditable record of plan, evidence sets and test for about
twice the tokens (Appendix~\ref{app:cost}).

Agents that write and run their own analysis help the open models (without
the codebook, EHRAgent gains 0.07, 0.24 and 0.10 over the single query for
the Qwen2.5 pair, qwen3.8 and gemma4) and, in the one frontier run, EHRAgent
with GPT-5.6 loses 0.08. Given the codebook, EHRAgent with gemma4 reaches 0.913, above the single
query (0.857) and below \art{} (0.981).\footnote{EHRAgent with the codebook
added to its prompt, gemma4 at a 16k context, all 1{,}500 test hypotheses;
outputs are in the review copy (Section~\ref{sec:setup}). Table~\ref{tab:main}
keeps the agents in their released, no-codebook setup.} DiscoveryBench's agent scores below the single query with the same model
and fails to run with Qwen2.5-Coder,
whose code calls a library function that no longer exists. An agent writes
its own denominator and test, and nothing checks them against the
codebook.

\subsection{RQ2: the staged pipeline}
\label{sec:families}

\label{sec:modular}
The gain comes from the contract and the verify-and-repair loop together. In
Table~\ref{tab:stage}, with the mapping withheld and the codebook given,
the verdict is right on 0.915--0.992 of the hypotheses when only S1 is
predicted, and the gemma4 planner matches the frontier; a plain prompt with the same codebook
and plan schema scores 0.86--0.92 with or without chain-of-thought, below \art{}'s planner for every open model (0.95, 0.92 and 0.99). Retrieval
is where the Qwen2.5 pair loses: Qwen2.5-Coder drops a sixth of the
hypotheses even with the gold plan. With the gold plan, qwen3.8 and gemma4 retrieve well, so their remaining loss is the plan. The open models do not compute the test reliably:
under the StatQA protocol on the gold counts they are wrong on a fifth to
two fifths of hypotheses, and the model's verdict on \art{}'s own counts loses
0.2--0.3 (Table~\ref{tab:main}). The \emph{w/o contract} row hands the plan to S2 as prose, removing only
the resolution of the named primitive into its rule; this costs the Qwen2.5 pair
0.14 and qwen3.8 and gemma4 0.03--0.05, so the weaker the model, the more
of the gain the contract carries. Verify-and-repair helps qwen3.8 and
gemma4 almost entirely at the first re-ask ($+0.14$ and $+0.13$ at $T = 1$,
at most $+0.01$ after) and costs the frontier models 0.01--0.02, where the
rechecks flag correct $n = k$ statements (Figure~\ref{fig:panels}c).

\art{}'s remaining errors concentrate in a few hypothesis families.
Figure~\ref{fig:families} breaks its evidence down by family. Overclaim
stays below 0.04 in every family except first/last, where qwen3.8 decides 0.20 of the undecidable hypotheses, and the Qwen2.5 pair and GPT-5.6 each decide 0.07. Each dark cell has a cause the plan exposes. The Qwen2.5 pair retrieves the wrong
set on first/last (evidence error 0.78) and abstains on 0.80 of the decisive
first/last hypotheses; the frontier models underclaim on threshold. One
failure does not show in the verdicts: gemma4 is right on 0.97 of proximity
hypotheses, but nearly a third of those right verdicts rest on a denominator
that admits non-adjacent pairs. Verdict accuracy follows the same pattern (Figure~\ref{fig:panels}a);
gemma4 scores at least 0.95 on every family. A second failure escapes every evidence panel: qwen3.8
scores 0.69 on position with every evidence set exact, because on 28 of the
30 negated position hypotheses its plan keeps the tail \emph{greater} after
marking the hypothesis negated; the test runs in the wrong direction and
supported and refuted swap (24 and 4). No plan check compares the tail with the negation.

Rewording the test hypotheses in the three other registers costs the open
models up to 0.13 under \art{} (Figure~\ref{fig:panels}b), and the planner
alone loses the same amount for qwen3.8 and gemma4
(Appendix~\ref{app:registers}), so paraphrase is paid at planning.

Figure~\ref{fig:tiers} holds the hypothesis shape fixed. Under \art{} every
model verifies the Easy and Medium hypotheses, and the open models lose 0.15 (Qwen2.5 pair), 0.10 (qwen3.8) and 0.01
(gemma4) when a second cohort adds a
denominator and a two-arm test. EHRSQL-C loses 0.68, 0.48 and 0.17 over the same steps, while the
frontier models stay near 1.0 in both systems. Table~\ref{tab:report} gives
the 180 single-report hypotheses, where S3 reads the answer off one timeline, with no test: every model but the Qwen2.5 pair scores 0.98--1.00 once given the
codebook, the pair reads a named report right on 0.43 of them without it, and
only \art{} brings the pair to 1.000.

\begin{table}[t]
\centering
\footnotesize
\setlength{\tabcolsep}{4pt}
\renewcommand{\arraystretch}{0.95}
\begin{tabular}{@{}llrrrrr@{}}
\toprule
 & & Qwen2.5 & qwen3.8 & gemma4 & Opus & GPT \\
Metric & Methods & 32B pair & 27B & 31B & 5.5 & 5.6 \\
\midrule
Acc & EHRSQL & .528 & .593 & .634 & .867 & .872 \\
 & EHRSQL-C & .577 & .744 & .857 & \textbf{1.000} & \textbf{.984} \\
 & \textbf{\art{}} & \textbf{.806} & \textbf{.869} & \textbf{.981} & .991 & .981 \\
\midrule
Ev.\ F1 & EHRSQL & .547 & .588 & .692 & .880 & .899 \\
 & EHRSQL-C & .643 & .751 & .836 & \textbf{1.000} & \textbf{1.000} \\
 & \textbf{\art{}} & \textbf{.789} & \textbf{.914} & \textbf{.975} & .998 & .993 \\
\midrule
Faithful & EHRSQL & .263 & .408 & .508 & .867 & .912 \\
 & EHRSQL-C & .405 & .697 & .861 & \textbf{1.000} & \textbf{1.000} \\
 & \textbf{\art{}} & \textbf{.864} & \textbf{.972} & \textbf{.945} & .998 & .999 \\
\midrule
Trace & EHRSQL & .143 & .217 & .305 & .751 & .786 \\
 & EHRSQL-C & .211 & .516 & .740 & \textbf{1.000} & \textbf{.979} \\
 & \textbf{\art{}} & \textbf{.635} & \textbf{.819} & \textbf{.918} & .967 & .956 \\
\bottomrule
\end{tabular}
\caption{Evidence behind the verdicts on the test set, for the
end-to-end systems that return evidence sets (Section~\ref{sec:metrics}); Acc
repeats Table~\ref{tab:main}; Ev.\ F1: Evidence F1; Faithful: Faithfulness. Bold: best per metric and model.}
\label{tab:evidence-all}
\vspace{-25pt}
\end{table}

\begin{figure}[t]
\centering
\includegraphics[width=\columnwidth]{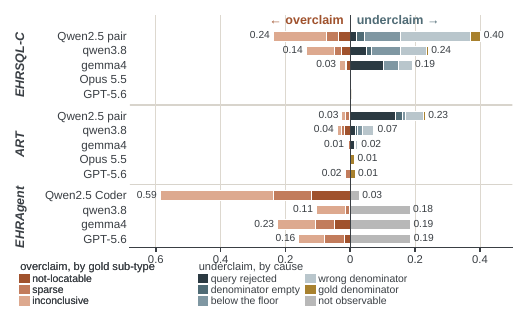}
\caption{The \emph{insufficient} class on the test set, by system and
model. Left: overclaim by gold sub-type. Right: underclaim by cause; grey
where the cause is not observable. Numbers are row totals.}
\label{fig:insufficient}
\vspace{-15pt}
\end{figure}

\subsection{RQ3: the verdict and its evidence}
\label{sec:evidence}
\label{sec:insufficient}

Every single-query and \art{} prediction records the reports it counted, so
each of their verdicts can be checked against the gold sets
(Table~\ref{tab:evidence-all}); the agents return no evidence sets, so they
have no row. End to end,
the verdict overstates what a system computed: the verdict--evidence gap.
With the Qwen2.5 pair, only 0.41 of EHRSQL-C's right verdicts rest on the
gold denominator, so 0.23 of its 0.577 accuracy does; its trace is 0.21 (Figure~\ref{fig:example}, panel~2).
Without the codebook the gap is wider: EHRSQL's denominators match the gold
on 0.26--0.91 of its right verdicts across the models. \art{} raises
faithfulness and trace for every open model, most for the weakest model,
and raises trace by more than accuracy. At the frontier EHRSQL-C and \art{} are nearly exact on every measure.

The \emph{insufficient} class shows the gap from the other side: a system
should decide exactly the hypotheses the data can decide.
Figure~\ref{fig:insufficient} splits overclaim by gold sub-type and
underclaim by cause, separating execution failure (a rejected query, an empty
arm) from retrieval error (a wrong denominator or one below the floor). Overclaim
comes mostly from the agents and the single-query baselines. EHRAgent with Qwen2.5-Coder decides 0.59 of the undecidable hypotheses,
most of them \emph{inconclusive}, because it runs its own test. EHRSQL-C overclaims mostly on \emph{inconclusive} hypotheses, where a wrong
denominator moves the tail, and fills \emph{not-locatable} denominators
with reports that do not qualify. \art{} overclaims on at most 0.04 of the undecidable hypotheses overall, because the harness test, not
the model, decides when a hypothesis is insufficient.

Underclaim is concentrated in the open models. EHRSQL-C with the Qwen2.5 pair answers \emph{insufficient} on nearly four decisive hypotheses in
ten, mostly on a wrong denominator or one below the floor. Under \art{}, the Qwen2.5 pair's remaining ones are mostly rejected
queries. The frontier models' few rest on the gold denominator, so the loss is in
the plan or the test, not in retrieval.

\section{Conclusions and Future Work}
\label{sec:conclusion}\label{sec:limitations}

We present \sys{}, a benchmark for temporal hypothesis verification
with gold for each decision: the plan, the evidence and the verdict.
Across five model configurations, systems fail mostly at choosing which records count,
a verdict--evidence gap that accuracy alone conceals. \art{} exposes each
decision as an inspectable output, closes most of this gap and brings a 31B
open model within 0.01 of the frontier models under the same paradigm. Benchmarks that ask language
models to test hypotheses against data should release the rules behind their
labels and score the evidence along with the answer.

The design points to its next steps. The codebook and the denominator
primitives are the only corpus-specific parts of \sys{}, so another corpus
of timelines enters by writing those two. Clinician-written hypotheses and
held-out templates would test whether systems carry the stated rules to
wording the templates never produced, and alternative codebooks would show
how far each verdict depends on the codebook chosen. Running the
codebook-given agents with every model would complete the matched
comparison.

\newpage
\bibliographystyle{ACM-Reference-Format}
\bibliography{related_work}

\appendix
\section{Positioning Against Related Work}
\label{app:positioning}

Table~\ref{tab:related-concise} sets the closest benchmarks and systems
against \sys{} along the decisions of Section~\ref{sec:task}.

\begin{table*}[htbp]
\centering
\scriptsize
\setlength{\tabcolsep}{4pt}
\renewcommand{\arraystretch}{1.25}
\begin{tabularx}{\textwidth}{@{}p{2.5cm} Y Y Y Y Y Y@{}}
\toprule
Work & Null & Test selection & Test execution & Temporal hypotheses & Clinical data & Gold released \\
\midrule
\multicolumn{7}{@{}l}{\textit{Benchmarks}} \\
StatQA \cite{zhu2024statqa} & \cpart{applicability only} & \cyes{model picks from a catalogue, scored} & \cno{none run} & \cno{no} & \cno{no} & \cno{final answer} \\
StatFormBench \cite{wang2026statformbench} & \cpart{test family} & \cpart{model names the family} & \cno{none run} & \cno{no} & \cno{textbook} & \cpart{formulation} \\
StatABench \cite{zhu2026statabench} & \cpart{test choice} & \cyes{model picks, scored} & \cpart{fixed tool functions} & \cno{no} & \cno{no} & \cno{final answer} \\
P-Bench / Fisher-R1 \cite{miao2026fisher} & \cno{given} & \cpart{model picks, unscored} & \cpart{model-written R} & \cpart{survival only} & \cpart{partly} & \cpart{reference $p$} \\
DiscoveryBench / BLADE \cite{majumder2025discoverybench,gu2024blade} & \cpart{open-ended} & \cpart{agent, rubric-scored} & \cpart{agent code} & \cno{no} & \cno{no} & \cpart{rubric} \\
EHRSQL \cite{lee2022ehrsql} & \cno{none (QA)} & \cno{none} & \cno{SQL; no test} & \cno{no} & \cyes{EHR} & \cpart{SQL, answer, abstention} \\
\midrule
\multicolumn{7}{@{}l}{\textit{Systems}} \\
EHRAgent \cite{shi2024ehragent} & \cno{none (QA)} & \cno{none} & \cno{agent code; no test} & \cno{no} & \cyes{EHR} & \cna{system} \\
LATCH \cite{gim2026latch} & \cno{given} & \cno{given by the source study} & \cyes{deterministic R pipelines} & \cno{no} & \cyes{NHANES} & \cna{system} \\
VERITAS \cite{veritas2026} & \cpart{planned by agent} & \cpart{agent plans} & \cpart{own code + rule} & \cno{no} & \cyes{imaging} & \cna{system} \\
\midrule
\textbf{\sys{} + \art{} (this work)} & \cyes{\textbf{composed}: kind, $p_0$, population} & \cyes{\textbf{derived from the plan} (null, arms); scored against the gold plan} & \cyes{\textbf{harness}: exact tests, floor} & \cyes{\textbf{7 families}} & \cyes{\textbf{VAERS timelines}} & \cyes{\textbf{plan, evidence, $p$, verdict}} \\
\bottomrule
\end{tabularx}
\caption{Positioning against related work by what each work asks and scores. \emph{Null}: who states the null and the
population it is tested over. \emph{Test selection}: who chooses the test
and whether that choice is scored. \emph{Test execution}: what computes the
statistic the verdict rests on, and with which tests. \emph{Temporal
hypotheses}: hypotheses about the order or timing of events within a report.
\emph{Gold released}: what a benchmark scores besides the final answer;
systems release no gold of their own. Marks: \cmark{} has it, \pmark{} partly,
\xmark{} lacks it, \nmark{} not applicable.}
\label{tab:related-concise}
\end{table*}

\section{Hypothesis Catalogue}
\label{app:stats}
\label{app:catalogue}

Table~\ref{tab:catalogue} lists all 26 template groups: the hypothesis as it is
worded in the \textsc{Original} register, which reports count toward its
denominator, the null, the test and the size of the released pool. Each
template also exists in the three other registers. Every cohort hypothesis
is tested one-sided at $\alpha = 0.05$ in each direction. The floor is
$n_{\min} = 10$ reports per arm: with $n_j = |D_j|$ the size of arm $j$, the
verdict is \emph{sparse} when $0 < \min_j n_j < 10$ and \emph{not-locatable}
when $\min_j n_j = 0$.

\begin{table*}[htbp]
\centering
\scriptsize
\setlength{\tabcolsep}{3.5pt}
\renewcommand{\arraystretch}{1.12}
\begin{tabularx}{\textwidth}{@{}l X p{3.1cm} p{1.6cm} p{2.25cm} rr@{}}
\toprule
Group & Hypothesis template (\textsc{Original} register) & Reports counted & Null & Test & Hyp. & Med.\ $n$ \\
\midrule
\multicolumn{7}{@{}l}{\textit{Easy: read off one report}} \\
P-R1 & In report $r$, $A$ occurred before $B$. & the named report & --- & none & 400 & --- \\
P-R2 & In report $r$, $A$ came first, before $B$. & the named report & --- & none & 400 & --- \\
C-R1 & In report $r$, $A$ and $B$ co-occur. & the named report & --- & none & 400 & --- \\
S-R1 & In report $r$, $A$, $B$, $C$ occur in the order $A$ $\rightarrow$ $B$ $\rightarrow$ $C$. & the named report & --- & none & 400 & --- \\
F-R1 & In report $r$, $A$ occurs first among $A$, $B$, $C$. & the named report & --- & none & 400 & --- \\
F-R2 & In report $r$, $A$ occurs last among $A$, $B$, $C$. & the named report & --- & none & 400 & --- \\
F-R3 & In report $r$, $A$ is the first of $A$, $B$, $C$ to occur. & the named report & --- & none & 400 & --- \\
T-R1 & In report $r$, $A$ occurred in the first half of the timeline. & the named report & --- & none & 400 & --- \\
Y-R1 & In report $r$, $A$ occurred in the stage immediately before $B$. & the named report & --- & none & 400 & --- \\
\midrule
\multicolumn{7}{@{}l}{\textit{Medium: tested in one cohort}} \\
P-V1 & After $v$, $A$ typically occurs before $B$. & both events, at distinct stages & $0.5$ & binomial & 501 & 17 \\
P-V3 & After $v$, $A$ usually comes before $B$. & both events, at distinct stages & $0.5$ & binomial & 501 & 17 \\
P-V2 & After $v$, $B$ follows $A$ more often than it follows $C$. & both events, at distinct stages & two rates equal & Fisher$^{*}$ & 597 & 22 \\
P-V4 & After $v$, $A$ precedes $B$ more often than $C$ precedes $D$. & both events, at distinct stages & two rates equal & Fisher$^{*}$ & 599 & 24 \\
C-V1 & After $v$, $A$ and $B$ typically co-occur. & both events present & $0.5$ & binomial & 600 & 19 \\
C-V2 & After $v$, $A$ and $B$ occur together more often than their separate rates would predict. & whole cohort, split on presence of $A$ & two rates equal & Fisher & 600 & 229 \\
C-V3 & After $v$, co-occurrence is the dominant relation for $A$ and $B$. & both events present & runner-up & modal vs.\ runner-up & 522 & 23 \\
F-V1 & After $v$, $A$ is most commonly the first event among $A$, $B$, $C$. & all candidates present, a unique first & runner-up & modal vs.\ runner-up & 102 & 14 \\
F-V2 & After $v$, $A$ is most commonly the last event among $A$, $B$, $C$. & all candidates present, a unique last & runner-up & modal vs.\ runner-up & 138 & 14 \\
F-V3 & After $v$, $A$ is the most common first event among $A$, $B$, $C$. & all candidates present, a unique first & runner-up & modal vs.\ runner-up & 102 & 14 \\
T-V1 & After $v$, $A$ appears in the earlier half of the timeline more often than chance would place it there. & $A$ present, off the middle stage & data-derived & binomial & 102 & 145 \\
Y-V1 & After $v$, when $A$ and $B$ fall in consecutive stages, $A$ usually comes first. & both events, at consecutive stages & $0.5$ & binomial & 204 & 17 \\
Y-V2 & After $v$, in reports where $A$ and $B$ are in consecutive stages, $A$ comes first more often. & both events, at consecutive stages & $0.5$ & binomial & 204 & 17 \\
Y-V3 & After $v$, $A$ and $B$ fall in consecutive stages more often than chance would give. & both events, at distinct stages & data-derived & binomial & 105 & 26 \\
H-V1 & After $v$, $A$ tends to precede $B$ in at least two out of three cases where their order is recorded. & both events, at distinct stages & bar in text, $2/3$ & binomial & 105 & 16 \\
H-V5 & After $v$, $A$ tends to be the first of $A$, $B$ and $C$ in at least half of cases. & all candidates present, a unique first & bar in text, $0.5$ & binomial & 66 & 13.5 \\
\midrule
\multicolumn{7}{@{}l}{\textit{Hard: compared across two cohorts}} \\
P-M2 & $A$ precedes $B$ more often after $v_1$ than after $v_2$. & both events, at distinct stages & two rates equal & Fisher & 87 & 21.5 \\
\bottomrule
\end{tabularx}
\caption{Hypothesis catalogue: the 26 template groups by difficulty level.
$A$--$D$ are event terms, $v$ a vaccine cohort and $r$ a report; for example,
P-V1 is realised as ``After Pfizer, Headache typically occurs before
Fatigue''. \emph{Reports counted}: the denominator rule. \emph{Null}:
$0.5$ is the symmetric null; \emph{runner-up} tests the named candidate
against its strongest competitor; \emph{two rates equal} compares
two arms; \emph{data-derived} computes chance from the denominator reports
(a report's share of stages in the earlier half, or the share of its distinct-stage event pairs that are adjacent); a \emph{bar in text} is
read from the wording. $^{*}$A report in both arms has its two outcomes
exchanged (Appendix~\ref{app:testdetails}). \emph{Hyp.}: released hypotheses; \emph{Med.\ $n$}:
median denominator size. Single-report hypotheses are read off one timeline with no
test. Cohort groups ship near-equal thirds of the three verdicts, except
C-V3 and Y-V3 (0.27 and 0.24 per decisive class), whose stricter tests leave
more hypotheses insufficient.}
\label{tab:catalogue}
\end{table*}

\section{Statistical Tests}
\label{app:testdetails}

Position and proximity hypotheses have data-derived nulls. Each report has its own
chance rate: its share of stages in the earlier half for position, and for proximity the share
of its distinct-stage event pairs that are adjacent. Under the null the
numerator follows a Poisson-binomial over these rates. The gold uses a
one-sided binomial at their mean, so the test runs from the counts and
one released rate, like every other test. The binomial is conservative
whenever its rejection cutoff lies at least one count beyond the expected
count \cite{hoeffding1956}, which holds in both tails for every such hypothesis in
the test set: its size is at most 0.039 per direction. Of the 180 test hypotheses with a
data-derived null (position and proximity, 90 each), the binomial leaves 74
insufficient; the Poisson-binomial would decide 9 of those 74 (one position
hypothesis and four proximity mirror pairs).

Argmax hypotheses (first/last and C-V3) use an exact binomial of
the named cell against its strongest observed competitor. In the hypothesis's
direction this is an intersection--union test against every competitor,
with size at most $\alpha$ \cite{berger1982}, because the binomial
$p$-value grows with the competitor's count. In the opposite direction any
competitor may win, so that tail is Bonferroni-adjusted over the
competitors. Both directions have size at most 0.049 on the test hypotheses.

In the two-arm hypotheses with shared reports (P-V2, P-V4), a report that orders
both pairs sits in both arms. The test exchanges its two outcomes, so the
report is counted once. Its exact null is the hypergeometric of the
unshared reports convolved with a fair binomial over the shared reports
whose outcomes differ, which is Fisher's test when no report is shared and
the exact McNemar test \cite{mcnemar1947} when every report is.

The codebook also lists the
asymptotic two-proportion test as valid for a two-arm hypothesis when every
expected cell is at least 5. A plan may choose it; the harness runs it only
when the arms share no report, and the gold always uses the exact two-arm
test above.

\section{Paraphrase Registers}
\label{app:registers}

Figure~\ref{fig:panels}b gives \art{}'s verdict accuracy in each paraphrase
register for the three open models; it is the data behind the register result
in Section~\ref{sec:families}. The planner alone, with retrieval and test at gold,
loses 0.10 (Qwen2.5 pair), 0.12 (qwen3.8) and 0.13 (gemma4) on the \textsc{Research}
register relative to \textsc{Original}, against 0.04, 0.11 and 0.13 for the
full pipeline.

\section{Harness and Release Details}
\label{app:harness}

\textit{\textbf{Sampling}.} Within each cohort group a seeded draw caps the
pool at 600 hypotheses, with each decisive core's hypothesis and mirror kept
together; the development set is disjoint from the test set on \texttt{fact\_id}.

\textit{\textbf{Validation screens}.} Under the permutation null matched
to each hypothesis shape, every group's share of decisive cores falls to about
$\alpha$; for position hypotheses it is 5.1--5.3\% of the cores, against
14--15\% under a naive null of 0.5.

\textit{\textbf{S1 prompt}.} The codebook is stated by situation, for
example ``a report with two events tied for earliest has no answer and is
excluded'' and ``order words are distance-blind; only wording that says the
two are adjacent conditions on consecutive stages''. The prompt also carries
a computed dataset card, the null sources and the test definitions. Summed
over the three open models, 84 test-set plans name no known hypothesis type;
these hypotheses are scored \emph{insufficient}, the gold verdict on 29 of them.

\textit{\textbf{Contract and checks}.} The contract holds one
conditioning rule per hypothesis type (about 280 lines of code) and a reverse
lookup from plan fields to hypothesis type with about ten normalisation rules.
The S1 loop runs 14 plan checks (a quantity against its denominator, a null
kind against its situation, a stated bar against the wording, a valid test
family, among others). The S2 loop runs three row checks (reports whose two
events co-exist in an order denominator, ties in a first/last denominator,
and non-consecutive pairs in a proximity denominator), plus two heuristic
rechecks: an arm whose denominator equals its numerator, and an empty arm
beside a full one.

\textit{\textbf{Intervals and paired tests}.} Accuracy intervals are 95\%
percentile bootstrap intervals that resample \texttt{fact\_id} clusters
(2{,}000 resamples), since a hypothesis and its mirror share their evidence.
Paired comparisons flip the sign of each fact's summed difference (10{,}000
permutations). On the test set, \art{} minus EHRSQL-C is $+0.229$ (Qwen2.5
pair), $+0.125$ (qwen3.8) and $+0.124$ (gemma4), each $p < 0.001$;
$-0.009$ for Opus 5.5 ($p < 0.001$) and $-0.003$ for GPT-5.6 ($p = 0.70$).

\textit{\textbf{Runtime}.} Open models run as 4-bit (Q4\_K\_M) GGUF weights
under Ollama 0.40.4 on one RTX A5000 (24 GB) with an 8{,}192-token context
and thinking off; the Qwen2.5 pair swaps its two models on the same GPU.
Frontier models run over their providers' APIs with prompt caching.

\section{Cost and Latency}
\label{app:cost}

Table~\ref{tab:cost} gives, per test hypothesis, the wall time of the sequential
harness, the tokens exchanged and the API cost from the providers' usage
counts at list price with prompt caching; local rows run on one GPU and have
no API cost. \art{} makes two base model calls per hypothesis (S1 and S2), plus up to
two re-asks per stage; EHRSQL-C makes one call. The table reports the two
base calls. With repair, the open models re-ask on most hypotheses (3.6--3.9
calls per hypothesis on average, at most 6) and exchange about twice the
tokens; the frontier models re-ask on 9\% of hypotheses and exchange 8\% more
tokens. All API runs on the test set and the single-report study (\art{} at
re-ask caps $k = 0$--$3$, EHRSQL, EHRSQL-C, StatQA and the stage rows) cost \$82 with Opus 5.5
and \$90 with GPT-5.6. GPT-5.6's hidden reasoning tokens (2.5M) fall mainly
on the rows where the model judges the test: 1.2M in the StatQA row and 0.9M
in EHRSQL and EHRSQL-C, against 0.3M in \art{}, whose test stage is
deterministic. EHRAgent with GPT-5.6 cost \$56 (4{,}501 calls, 11.3M prompt tokens of which
5.0M cached, 1.4M completion tokens, reasoning off). A 31B open model answers
a test hypothesis in 14~seconds on one workstation GPU and a 27B one in 10.

\begin{table}[htbp]
\centering
\footnotesize
\setlength{\tabcolsep}{3pt}
\begin{tabular}{llrrr}
\toprule
System & Model & Seconds & Tokens & API cost (USD) \\
\midrule
\art{} & Opus 5.5 & 7.1 & 9{,}150 & 0.014 \\
\art{} & GPT-5.6 & 6.2 & 5{,}600 & 0.012 \\
\art{} & Qwen2.5-32B pair & 12.3 & 5{,}000 & 0 \\
\art{} & qwen3.8-27B & 10.1 & 5{,}150 & 0 \\
\art{} & gemma4-31B & 14.1 & 5{,}300 & 0 \\
EHRSQL-C & Opus 5.5 & 5.6 & 4{,}850 & 0.013 \\
EHRSQL-C & GPT-5.6 & 8.0 & 3{,}200 & 0.013 \\
EHRSQL-C & Qwen2.5-Coder-32B & 2.8 & 2{,}850 & 0 \\
EHRSQL-C & qwen3.8-27B & 6.2 & 3{,}050 & 0 \\
EHRSQL-C & gemma4-31B & 6.1 & 3{,}100 & 0 \\
\bottomrule
\end{tabular}
\caption{Cost and latency per hypothesis (test set for the local models, usage
counts from the test runs for the frontier models). \art{} rows are the two base calls without repair; tokens are prompt plus
completion, and GPT-5.6's completion count includes its hidden reasoning
tokens. Local latencies are measured on one RTX A5000 with the
model fully resident; the Qwen2.5 pair's latency includes swapping the two models
between stages.}
\label{tab:cost}
\end{table}

\clearpage

\end{document}